\documentclass[conference]{IEEEtran}
\IEEEoverridecommandlockouts
\usepackage{stfloats}
\usepackage{cite}
\usepackage{graphicx}
\usepackage{subfig}
\usepackage{amsmath,amssymb,amsfonts}
\usepackage{algorithmic}
\usepackage{graphicx}
\usepackage{textcomp}
\usepackage{xcolor}
\begin{document}

\title{Early Prediction of Acute Kidney Injury in Critical Care Setting Using Clinical Notes\\
\thanks{This work was supported in part by NIH Grant 1R21LM012618.}}

%
%

\author{
\IEEEauthorblockN{Yikuan Li}
\IEEEauthorblockA{\textit{Dept. of EECS } \\
\textit{Northwestern University}\\
Evanston, IL, U.S.A \\
yikuanli2018@u.northwestern.edu}
\and
\IEEEauthorblockN{Liang Yao}
\IEEEauthorblockA{\textit{Dept. of Preventive Medicine } \\
\textit{Northwestern University}\\
Chicago, IL, U.S.A \\
liang.yao@northwestern.edu}
\and
\IEEEauthorblockN{Chengsheng Mao}
\IEEEauthorblockA{\textit{Dept. of Preventive Medicine } \\
\textit{Northwestern University}\\
Chicago, IL, U.S.A \\
chengsheng.mao@northwestern.edu}
\and
\IEEEauthorblockN{Anand Srivastava}
\IEEEauthorblockA{\textit{Div. of Nephrology and Hypertension} \\
\textit{Northwestern University}\\
Chicago, IL, U.S.A \\
anand.srivastava@northwestern.edu}
\and
\IEEEauthorblockN{Xiaoqian Jiang}
\IEEEauthorblockA{\textit{School of Biomedical Informatics} \\
\textit{Univ. of Texas Health Science Center}\\
Houston, TX, U.S.A \\
xiaoqian.jiang@uth.tmc.edu}
\and 
\IEEEauthorblockN{Yuan Luo (corresponding)}
\IEEEauthorblockA{\textit{Dept. of Preventive Medicine} \\
\textit{Northwestern University}\\
Chicago, IL, U.S.A \\
yuan.luo@northwestern.edu}
}

%
%
\maketitle

\begin{abstract}
Acute kidney injury (AKI) in critically ill patients is associated with significant morbidity and mortality. Development of novel methods to identify patients with AKI earlier will allow for testing of novel strategies to prevent or reduce the complications of AKI. We developed data-driven prediction models to estimate the risk of new AKI onset.  We generated models from clinical notes within the first 24 hours following intensive care unit (ICU) admission extracted from Medical Information Mart for Intensive Care III (MIMIC-III). From the clinical notes, we generated clinically meaningful word and concept representations and embeddings, respectively. Five  supervised learning classifiers and knowledge-guided deep learning architecture were used to construct prediction models. The best configuration yielded a competitive AUC of 0.779. Our work  suggests that natural language processing of clinical notes can be applied to assist clinicians in identifying the risk of incident AKI onset in critically ill patients upon admission to the ICU. 
\end{abstract}

\section{Background}
Acute kidney injury (AKI) is commonly encountered in adults in the intensive care unit (ICU) \cite{ali2007incidence}. Patients with AKI are at risk for adverse clinical outcomes such as prolonged ICU and hospitalization stays, need for renal replacement therapy, future development of chronic kidney disease (CKD), and increased mortality \cite{kellum2013diagnosis}. Unfortunately, serum creatinine (SCr), which is used to identify changes in renal function, is a late marker of injury \cite{de2012biomarkers}. However, the efficacy of intervention greatly relies on the early identification of deterioration \cite{kellum2013diagnosis}. It is critical to have an early recognition in that AKI usually occurs over the course of a few hours to days and is potentially reversible if detected and managed timely \cite{li2013acute}.

Prior risk prediction models for AKI based on EHR data yielded modest performance as evaluated by AUC around 0.75. These prediction models usually are limited to either a certain set of biomarkers or a specific group of patients. For example, many studies focus on the validation of novel biomarkers or risk scores \cite{perazella2015urine}. Others were designed to predict AKI risks for specific patient groups, such as cardiac surgery patients \cite{kim2013simplified}, elderly patients \cite{kate2016prediction} and  pediatrics patients \cite{sanchez2016development}.

Epidemiology studies show multiple comorbidities, including diabetes mellitus, cardiovascular disease, chronic liver disease, cancer, and complex surgery have been associated with the development of AKI \cite{bellomo2012acute}. Only a comprehensive understanding of ICU patients can help us forecast the risk for development of AKI. To this end, We address the problem by developing AKI predictors with careful modeling of the content from clinical notes of ICU patients within the first 24 hours of admission using natural language processing. To the best of our knowledge, this is the first study that uses clinical notes to predict early AKI onset for general adult patients in critical care setting. Our work demonstrates that effectively applying NLP to clinical notes and extracting meaningful features can lay the foundation of building machine learning models that are predictive for AKI onset risk early on. From a practical point of view, our prediction model could be used to alert clinicians of critically ill patients at high risk for developing AKI soon after ICU admission. 

\section{Methods}
\begin{figure}
\begin{center}
\includegraphics[width=0.9\linewidth,height =0.95 \linewidth]{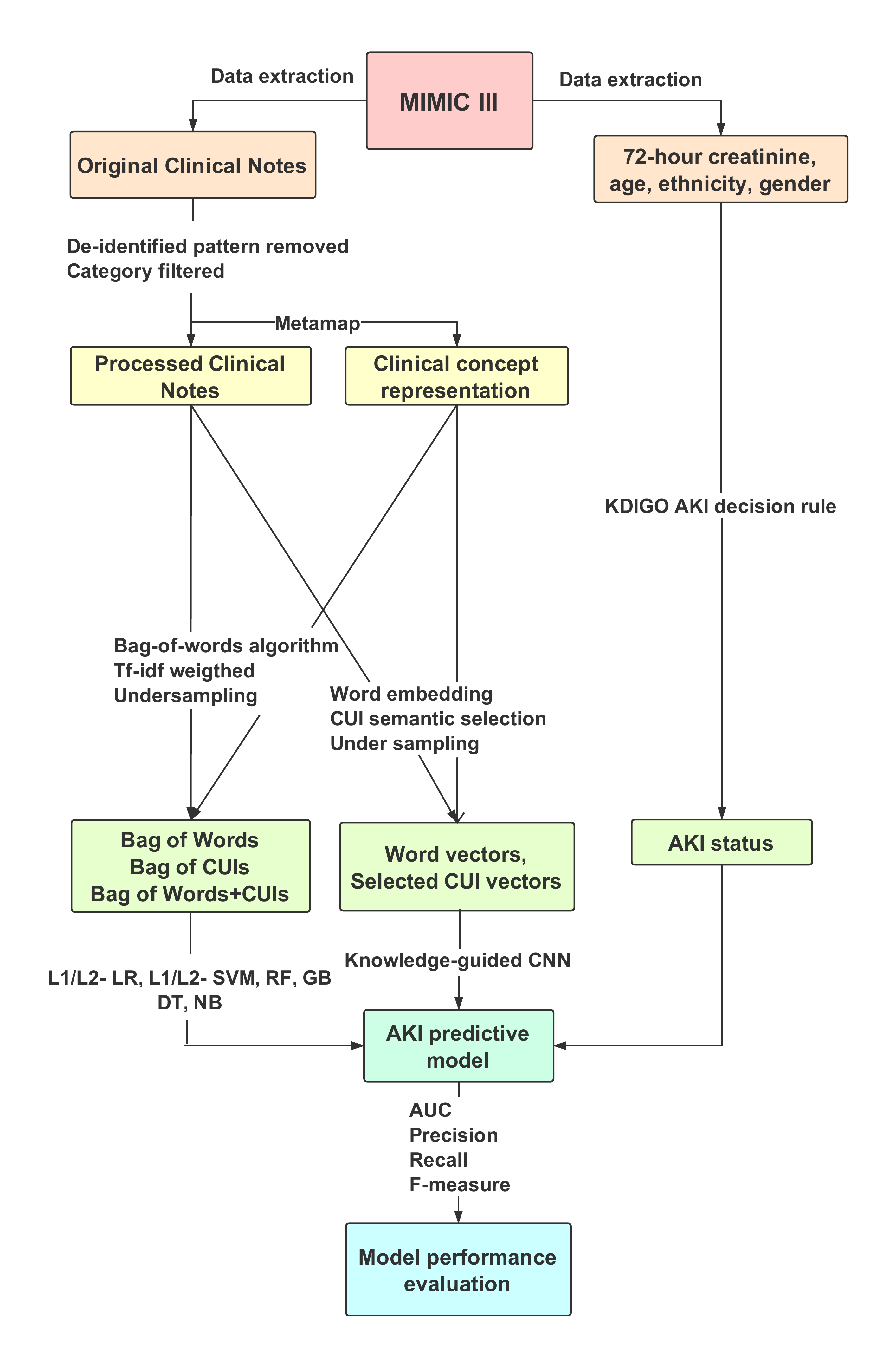}
\end{center}
\caption{The study design. We extracted patient demographics, clinical notes during the first day of ICU admission and 72-hour serum creatinine after admission from MIMIC III database. KDIGO AKI criterion was used to obtain AKI status. We applied and combined different data representation methods, imbalanced learning strategies, and supervised learning algorithms as well as deep learning architectures to build classifiers. F1 score, precision, recall, and AUC were used to evaluate the model performance.}
\label{fig:Diagram}
\end{figure}

Fig. \ref{fig:Diagram} shows the overall flowchart of the study design, with details explained in the following sections.
\subsection{Dataset}
Data for this study was extracted from the freely accessible critical care database, Medical Information Mart for Intensive Care III (MIMIC-III) \cite{johnson2016mimic}. We only included patients who are 18 years or older. Data extracted included patient age, gender, race/ethnicity, clinical notes during the first 24 hours of ICU admission and 72-hour serum creatinine after admission. Kidney Disease Improving Global Outcomes (KDIGO) \cite{kidney2009kdigo} was adopted as the definition of AKI to generate the ground-truth label for our predictive model. In order to avoid unnecessary alert, patients who had AKI, history of Chronic Kidney Disease (CKD) on admission (Day 1), or whose clinical notes have mentioned kidney dysfunction related words and their abbreviations, were excluded. Multiple notes of one single ICU stay were concatenated together to form one record. To make sure patients are receiving active care, only ICU stays containing at least one physician note or nursing note were retained. 77,160 clinical notes of 14,470 patients' 16,560 ICU stays were included in our final dataset, of which the prevalence is 16.7\%. The entire dataset was split in approximate ratio 7:3 to training and held-out test sets. Multiple ICU stays of the same patient were assigned into either the training set or the held-out test set but not both.

\subsection{Clinical word and concept representation}
To interpret clinical notes better, some preprocessing steps are needed. Masked private health information (PHI) in the notes were removed first. Then, each word was stemmed using Porter stemming to reduce inflectional variations \cite{porter1980algorithm}.

To be used by machine learning classifiers, clinical notes should be converted to structured features. Our first set of features consists of unigram Bag-of-words, which identified and normalized lexical variants from the unstructured text content. Those features with document-frequency under 100 were removed to reduce noise. A total of 313 stop words were applied according to NCBI guide. Term frequency-inverse document frequency (tf-idf) weighting adjustment was also applied. MetaMap \cite{aronson2010overview} was used to first identify the medical concepts from free text clinical notes. The extracted concept features are the Concept Unique Identifiers (CUIs) from Unified Medical Language System (UMLS). Three feature sets, Bag-of-Words,  Bag-of-CUIs and Bag-of-Words+CUIs (the combination of above two sets), were prepared for downstream machine learning classifiers.

\subsection{Machine learning classifiers and Knowledged-guided CNN}
Five algorithms were used to classify the ICU stays and stratify the AKI risks, including multinomial na\"ive Bayes (NB) as baseline, L1- and L2-regularized support vector machine (SVM) with linear kernel, L1- and L2-regularized logistic regression (LR) and two ensemble algorithms, random forest (RF) and gradient boosting decision tree (GBDT). To down weight the majority class, random under-sampling (RUS) was involved to the experiment.

We also explored deep learning models for AKI prediction. We used Knowledge-guided Convolutional Neural Networks (CNN) to combine word features and UMLS CUI features \cite{yao2018clinical}. It used pre-trained word embeddings and CUIs embeddings of clinical notes as the input. We adopted the same architecture and parameter settings for Knowledge-guided CNN as in \cite{yao2018clinical}. To address imbalance, We experimented random under-sampling with the following training class ratio, 1:1, 1:2 and 1:3.

In order to tune parameters, reduce over-fitting and instability, 5-fold cross-validation was performed on training set. We adopted AUC to evaluate the performance of imbalanced binary classifiers. Precision, recall, F-measure of positive AKI status are also calculated for reference.

\section{results}
\begin{table*}[t]
\caption{Performance of five  supervised learning classifiers over three different clinical representation features, CNN over word embedding vectors and knowledge- guided CNN over CUI+word embedding vectors with random under-sampling methods applied to majority class}
\begin{center}
\begin{tabular*}{\linewidth}{@{\extracolsep{\fill}}c c c c c c c}
\hline
\textbf{Feature}&\textbf{Algorithm}&\textbf{Undersampling}&\textbf{AUC}&\textbf{Precision}&\textbf{Recall}&\textbf{F-measure}\\
\hline
\textbf{Bag of words}    &   NB  & N/A    &  0.7454&  0.6018  &  0.0793  &0.1401\\
&L2- SVM&N/A  &   0.7790&  0.4072&    0.6469&  0.4998\\
&L1- SVM&N/A  &  0.7730&  0.3765&    0.6564&  0.4769\\
&L2- LR& N/A  &  0.7790&  0.4068&    0.6539&  0.5016\\
&L1- LR& N/A  &   0.7715&  0.3835&    0.6504&  0.4825\\
&RF&  N/A   &     0.7659&  0.5227&    0.0268&  0.0509\\
&RF&  1:1 RUS   &     0.7613&  0.3892&    0.6520&  0.4612\\
&GBDT& N/A    &   0.7673&  0.6146&    0.0688&  0.1237\\
&GBDT& 1:1 RUS    &   0.7710&  0.3685&    0.6436&  0.4728\\
\textbf{Bag of CUIs}&        NB& N/A &        0.7448&  0.5263&    0.0699&  0.1235\\
&L2- SVM&N/A &    0.7684&  0.3964&    0.6329&  0.4874\\
&L1- SVM&N/A &    0.7688&  0.3770&    0.6504&  0.4773\\
&L2- LR&N/A &     0.7690&  0.3916&    0.6376&  0.4851\\
&L1- LR&N/A &     0.7649&  0.3686&    0.6457&  0.4693\\
&RF&N/A &         0.7633&  0.4385&    0.5198&  0.4757\\
&RF&  1:1 RUS   &     0.7627&  0.3491&    0.6234&  0.4697\\
&GBDT&N/A &       0.7643&  0.5663&    0.0548&  0.0999\\
&GBDT& 1:1 RUS    &   0.7703&  0.3531&    0.6668&  0.4552\\
\textbf{Bag of words+CUIs}&  NB&N/A &         0.7448&  0.5263&    0.0699&  0.1235\\
&L2- SVM&N/A &    0.7785&  0.4089&    0.6410&  0.4993\\
&L1- SVM&N/A &    0.7749&  0.3813&    0.6457&  0.4795\\
&L2- LR&N/A &     \textbf{0.7791}&  0.4070&    0.6504&  0.5007\\
&L1- LR&N/A &     0.7730&  0.3863&    0.6492&  0.4844\\
&RF&N/A &         0.7676&  0.4839&    0.0175&  0.0338\\
&RF&  1:1 RUS   &     0.7652&  0.3468&    0.6532&  0.4713\\
&GBDT&N/A &       0.7671&  0.5892&    0.0886&  0.1540\\
&GBDT& 1:1 RUS    &   0.7701&  0.3414&    0.6641&  0.4637\\
\textbf{Word embeddings}&CNN&N/A &0.7269&0.5016&0.1818&0.2669\\
&CNN&1:1 RUS&0.7224&0.2848 &0.7040&0.4055\\
&CNN&1:2 RUS&0.7285&0.4364&0.4476&0.4419\\
&CNN&1:3 RUS&0.7381&0.5010&0.2949&0.3712\\
\textbf{Word embeddings + Semantic selectd CUIs}&Knowledge-guided CNN&N/A &0.7231&0.4437&0.2436&0.3145\\
&Knowledge-guided CNN&1:1 RUS&0.7466&0.3590&0.6294&0.4572\\
&Knowledge-guided CNN&1:2 RUS&0.7529&0.4349&0.5175&0.4726\\
&Knowledge-guided CNN&1:3 RUS&0.7341&0.4483&0.4347&0.4414\\
\hline
\end{tabular*}
\label{tab:results}
\end{center}
\end{table*}

Table \ref{tab:results} presents the results of different classifiers over three feature sets. We first compared the performance on three different features. Bag-of-words and the hybrid feature yielded quite similar performance with different classifiers and both slightly outperformed bag-of-CUIs as features. This is likely because when identifying the medical concepts with MetaMap, some useful medical concepts may not be recognized, while some spurious medical concepts may be falsely identified. In addition, the presence of concept ambiguity in MetaMap also poses challenge to downstream machine learning algorithms\cite{zeng2018natural,luo2016bridging}.

Next, we compared the performance of different classifiers. LR and SVM with linear kernel have slightly better AUCs than the two ensemble classifiers, RF and GBDT. The fact that decision tree models randomly samples subset of the training data might result in lost of key features for prediction, on the contrary, LR and SVM use entire training set. In addition, RF and GBDT over different features all yielded a very low recall less than 0.1 because data is imbalanced. Therefore, we resorted to random under-sampling to improve the performance of the two ensembles classifiers to a reasonable range.
For Knowledge-guided CNN all configurations of CNN architecture and embeddings did not outperform the non-CNN classifiers. Based on our previous experiments on 20 News, R52 and R8 of Reuters corpus, Movie Review, AG News etc. corpora and findings in \cite{zhang2015character}, CNN-based architectures generally work well for large scale data sets with short texts, while may not outperform Bag-of-word on smaller corpus with long texts like our AKI clinical note corpus.


\section{Discussion}
\subsection{Feature analysis}
\begin{figure}
  \centering
  \subfloat[Bag-of-Words]{\includegraphics[width=0.25\textwidth]{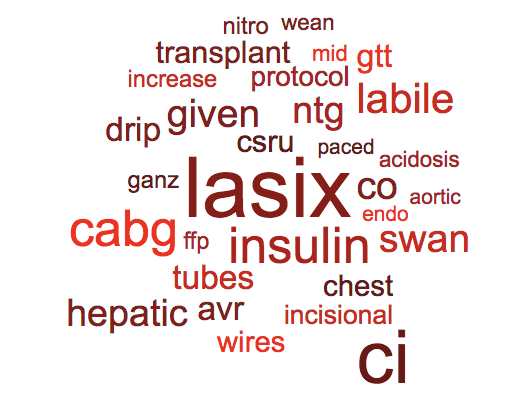}}
  \subfloat[Bag-of-CUIs]{\includegraphics[width=0.25\textwidth]{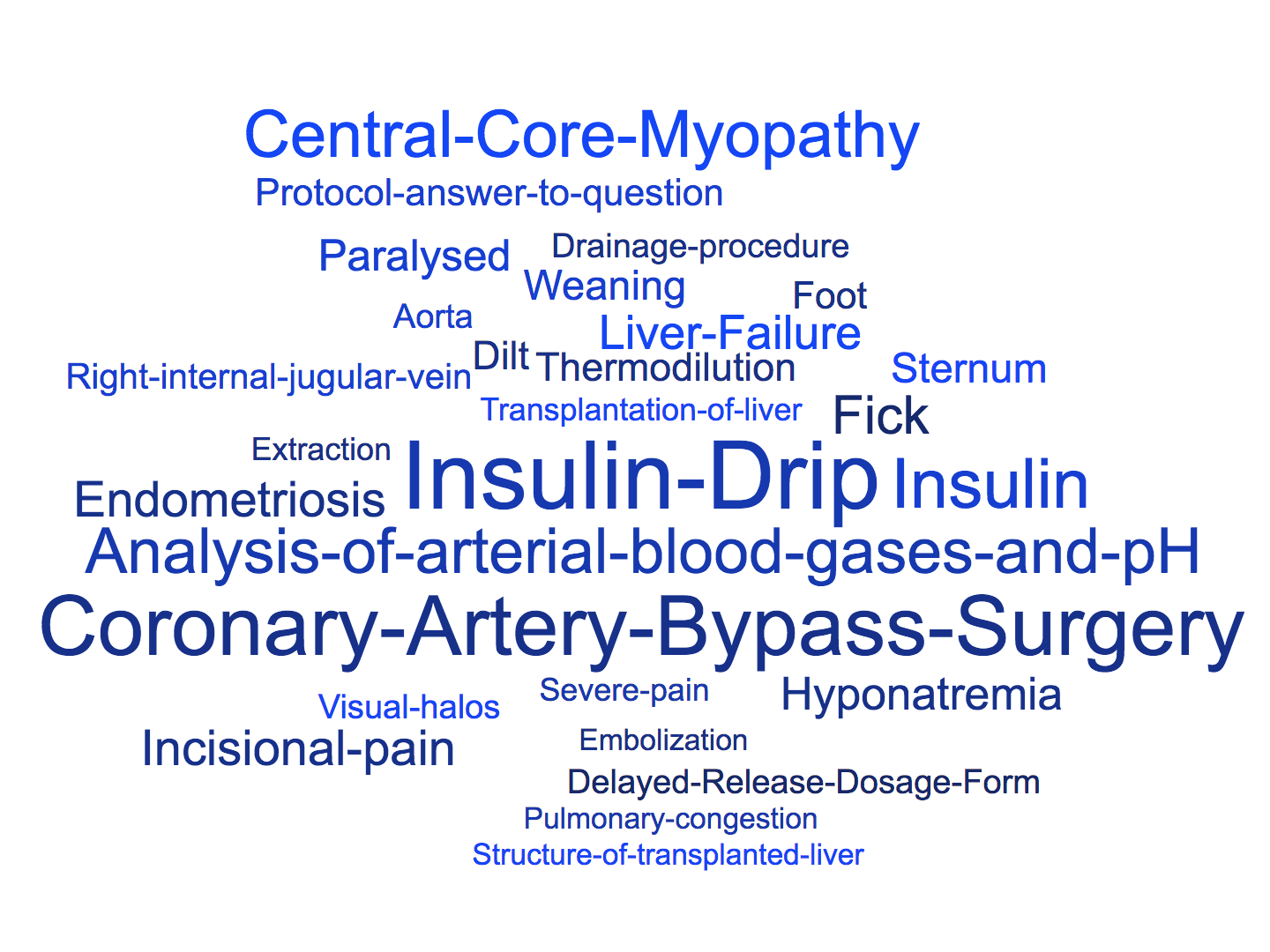}}  
  
  \caption{Ranked top 30 positive features in two feature sets with its coefficients as font size in AKI onset prediction. Model: L2-regularized logistic regression.}
   \label{fig:wc}
\end{figure}

We further examined important features by ranking coefficients in the L2-regularized logistic regression over bag-of-words and bag-of-CUIs as Fig. \ref{fig:wc}. In most case these features appear to be clinically meaningful. For example, \lq lasix' is one of diuretic that can treat fluid retention and swelling that might be caused by kidney dysfunction. The word \lq swan', which refers to \lq Swan-Ganz pulmonary artery catheter', is used to evaluate cardiac filling pressure and may help identify individuals who are in congestive heart failure\cite{mullens2009importance}. The word \lq cabg' (short for \lq Coronary artery bypass surgery') indicates an important clue since these individuals may develop kidney injury post-operatively. 'Central-core-myopathy' with highest coefficient in bag-of-CUIs, is a dominantly inherited congenital myopathy. Rhabdomyolysis is another form of myopathy that could lead to severe AKI \cite{quinlivan2003central}. Some features have high coefficients in both representations, e.g. \lq cabg', \lq insulin' and \lq incisional', present a high correlation with AKI onset. Feature examination confirms that clinically meaningful key words and key concept can be used to predict AKI onset and the models we built do capture those words and concepts.

\subsection{Error analysis}
In error analysis, we investigated 10 ICU stays in training set that are wrongly classified by L2- LR over Bag-of-Words with high probability. One patient was worngly predicted to AKI onset with 0.908 probability. In his first 24-hour notes, we found that he had just received CABG and chest tube removal. In his radiology findings part, pneumothorax and bibasilar pleural effusions were seen and a retrograde cardiac opacity was suspected. AKI was incorrectly predicted since this patient's notes contained many high coefficient words such as cabg, pneumothorax, and labile. Although he did not develop AKI, patients are at great risk after cardiac surgery. Our model should be tolerant to this kind of error.

In another case, one 41-year-old man was predicted to AKI onset with only 0.204 probability, while he developed AKI afterwards. He was admitted with diagnosis: hypoxia and decreased breath sounds. In his notes, we did not find high coefficient words written frequently. However, we found that he had a history of urosepsis. Septic shock from a urinary source leads to hypotension and subsequently decreased blood flow to major organs, which could lead to AKI  \cite{zarjou2011sepsis}. Only 79 out of 11560 patient notes in training set used the term urosepsis. Clinically, many people have urinary tract infections and then sepsis or septic shock from this source. The physicians or nurses may not always use the words 'urosepsis'. We could overcome this error by better terminology mapping.
\subsection{Limitations and Future Work}
There also some limitations of our study. First, we only adopted MetaMap to identify the medical concepts from free text clinical notes. We didn't examine other phenotyping system for comparison result. Additionally, we only investigated notes from MIMIC III database. To be generalizable, further work on more datasets is needed. Moreover, we only adopted Knowledge-guided CNN as deep learning framework. Searching for other deep learning architecture that is suitable for small scale dataset with long text content has the possibility of further improvement. Finally, we plan to perform additional validation on external patient cohorts and bring our model into practice as a clinical decision support system.

\section{Conclusion}
Our study demonstrates that a supervised learning-based NLP approach is useful to predict AKI onset early on in the first 72 hours following ICU admission for general adult patient population in the critical care setting. We showed that well represented clinical notes as features can predict new AKI onset as defined by KDIGO with competitive AUC of 0.779 by  supervised learning classifier, while previous studies focused on specific patient groups or on novel biomarkers. Further research is needed to prospectively validate the model and evaluate its clinical utility for identifying at risk patients early in their hospital course.
\bibliographystyle{IEEEtran}
\bibliography{reference}
\end{document}